# On Quantum Decision Trees

Subhash Kak

*Abstract*

Quantum decision systems are being increasingly considered for use in artificial intelligence applications. Classical and quantum nodes can be distinguished based on certain correlations in their states. This paper investigates some properties of the states obtained in a decision tree structure. How these correlations may be mapped to the decision tree is considered. Classical tree representations and approximations to quantum states are provided.

## INTRODUCTION

Imagine a decision system where the choices are made according to the measurements of a quantum state [1][2]. These choices may be mapped into a tree with each measurement is represented by a node (e.g. [3]). The node itself may be a human agent whose cognitions are modeled as a quantum system [4]-[7], or a classical agent that has access to quantum resources. Quantum models of cognition and information processing provide new insight into the workings of human agents in different decision environments (e.g. [8]-[11]) although it comes with features that do not have analogs in classical information [12][13].

Considerations under which quantum probability is the appropriate measure to use for cognitive processes have been examined from a variety of perspectives (e.g. [14]). In human agents, cognitive dissonance describes the stress arising out of holding two contradictory theories or ideas or actions [15][16]. The fact of such a state with contradictions is suggestive of superposition and thus it may justify the idea of quantum cognition. Quantum-like models have also been proposed to explain certain biological processes [17]. In engineered system the issue of dissonance is irrelevant and when considering quantum resources, mutually exclusive states will coexist.

There is much research that has gone into measurements that can distinguish between classical and quantum states as evidenced by the great interest in hidden variable theories and the violation of Bell's Inequalities [18][19]. The distinction between the two is on account of the fact that quantum states come with entanglement as a characteristic that shows up in nonlocality in physical processes and order effects in probability. The difference between classical and quantum domains can under certain conditions be ascertained by a condition on the ratios of the outcomes being the same ($P_s$) to being different ($P_n$) for 3-way coincidences. As shown earlier [20] for the classical and the maximally entangled quantum ($|\varphi\rangle = \frac{1}{\sqrt{2}}(|00\rangle + |11\rangle)$) this ratio is given by:

$$\frac{P_s}{P_n} = \begin{cases} \frac{1}{2}, \text{for classical} \\ \frac{1}{3}, \text{for quantum} \end{cases} \quad (1)$$

This means that to distinguish between the two in the presence of noise, the threshold of 5/12 (the mid-point) should be used. It was also shown [20] that for non-maximally entangled objects given by the state $|\varphi\rangle = \frac{1}{\sqrt{1+r^2}}(r|00\rangle + |11\rangle)$ the separation can be done so long as *r* < 5.83.

Order effect in quantum probability implies that P(AB) is not necessarily equal to B(BA). A similar effect characterizes responses of groups of people and survey designers know that the order in which questions are asked has a bearing on the response [21]. In cognitive choices, knowledge of a prior event may affect later choices. It has been suggested that the response depends on which schemas (of memories or attitudes) are most readily available in the mind and also on underlying quantum ground to the cognitive system.

In this paper we consider agents with quantum resources. We show how the difference between classical and quantum processes may by accounted for by varying branch-probabilities in the decision tree and that this is sufficient to explain probability order effects.

## JOINT PROBABILITY AND ENTANGLEMENT

Consider an urn with a large number of balls that are black, white, or half-black and half-white and the sampling protocol is to put the ball back into the urn after it is examined. If there are no half-black and half-white balls and the number of black and while balls is identical, this procedure will produce the same probabilities for the events (i) observed ball is white; and (ii) observed ball is black; or P(W) = P(B) = ½.

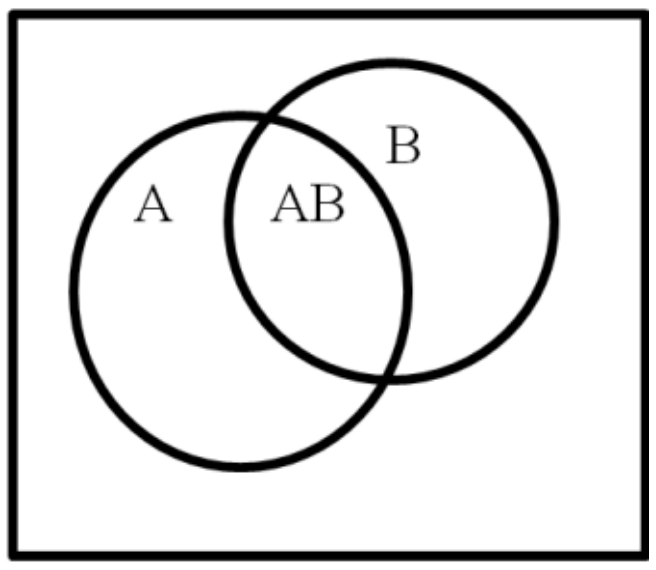

Figure 1. Venn diagram for classical probability where P(AB)=P(BA)

If sampling is done is sequence, the results for, say, two consecutive choices will be the same irrespective of the order of choices. In other words, P(WB) = P(BW) = ¼. If the urn has 60 purely black balls, 10 half-black and half-white balls, and 30 purely white balls, we will get P(B) = 0.7; P(W) = 0.4; P(BW) = P(WB) = 0.1. In general, for two events A and B belonging to the same sample space that are related to classical objects,

$$P(AB) = P(A|B)\,P(B) = P(B|A)\,P(A) = P(BA) \quad (2)$$

The probability of a specific joint event in a quantum state is the absolute value square of the corresponding probability amplitude and there are no constraints on what these amplitudes are. For another perspective on why equation (2) may not hold for a quantum state one must consider entanglement, which resource is absent in a classical state. The fact that quantum entanglement can be classically simulated [22] makes it worthwhile to look at the possible relationship between classical and quantum decision trees.

## DECISION TREES

Many interesting computational problems are most clearly formulated in terms of decision trees that are commonly used in operations research and in machine learning to help represent a strategy most likely to reach a goal. For simplicity, we consider binary trees. In such a tree, each node has at most two children. Figure 4 represents the case of a decision tree dealing with a game where the objective is to obtain two 1s and the payoff for the nodes that are successful is listed by A, B, and C, respectively. This tree could be classical or quantum.

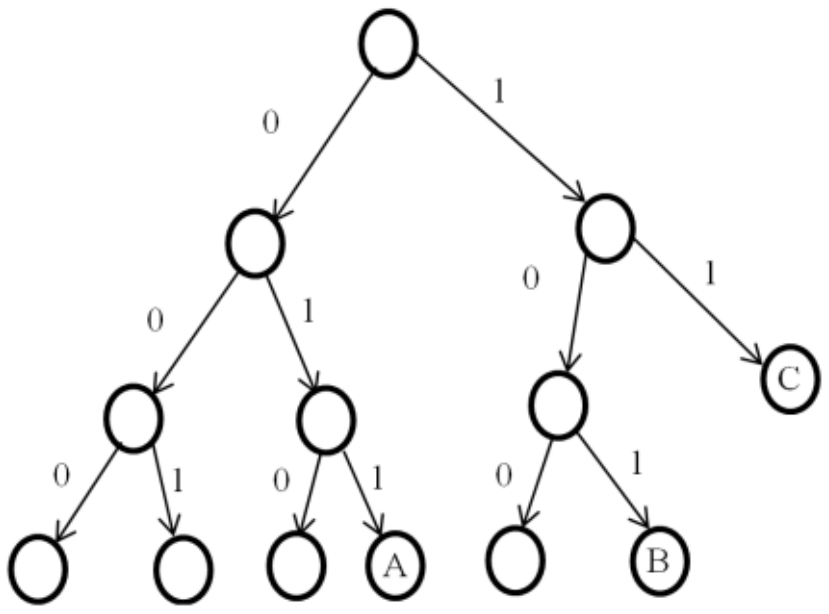

Figure 2. A binary decision tree

If the system nodes are indistinguishable, the probability of branching at each level can be taken to be fixed and let it be p for 0 and (1-p) for 1. The payoff for this tree will be $p(1-p)^2(A+B)+(1-p)^2C$ and one may easily determine the condition for which the payoff is maximized.

*Definition*. A decision tree with unchanging branching probabilities is a *fixed tree*.

*Definition*. A decision tree in which the branching probabilities may vary across nodes is a *random tree*.

We now state the following property which is quite obvious:

**Theorem 1**. A quantum state with arbitrary probability amplitudes may be mapped into a random tree.

*Reconstruction Procedure*. The following procedure establishes the idea behind the theorem. Suppose the quantum state is $|\varphi\rangle = a|00\rangle + b|01\rangle + c|10\rangle + d|11\rangle$. The sum of the probability of the events starting with 0 is $|a|^2 + |b|^2 = m$ and that of starting with 1 is $|c|^2 + |d|^2 = n = 1 - m$. Therefore, at the first level, the branching to 0 has probability of *m* and of 1 has probability *n*=1-*m*. At the second level, the corresponding probabilities will be $\frac{|a|^2}{m}$ and $\frac{|b|^2}{m}$. Likewise, at the second node of the second level, the branching probabilities are $\frac{|c|^2}{n}$ and $\frac{|d|^2}{n}$. This yields the correct probabilities of $|a|^2, |b|^2, |c|^2, \text{and } |d|^2$. This method can be applied to any number of levels of the tree.

*Example 1*. Consider

$$|\varphi\rangle = 0.6|000\rangle - 0.5|001\rangle + 0.2|010\rangle + 0.2|011\rangle + 0.3|100\rangle + 0.3|101\rangle + 0.2|110\rangle + 0.3|111\rangle \quad (3)$$

The probability of the various outcomes is 0.36, 0.25, 0.04, 0.04, 0.09, 0.09, 0.04, 0.09, respectively. Using the procedure outline above, we get the decision tree with the probabilities shown in Figure 3, where it is understood that the left branching is 0 and the right branching is 1.

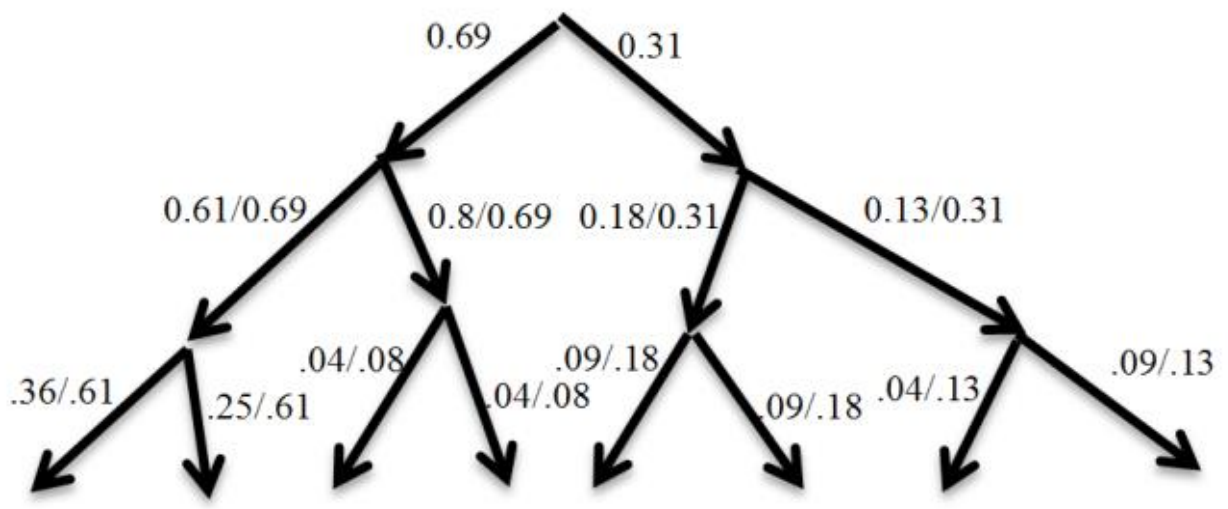

Figure 3. The decision tree for the quantum state (3)

Since quantum objects are in a superposition state, the outcome of sampling depends on the conditions under which the experiment is being conducted and the nature of the state. This is why the branching probabilities are not fixed. Consider a quantum system in terms of two independent qubits with the following density functions:

$$\rho_X = \begin{bmatrix} a & b \\ c & d \end{bmatrix} \text{ and } \rho_Y = \begin{bmatrix} e & f \\ g & h \end{bmatrix} \quad (4)$$

The joint density function of the two qubits will then be:

$$\rho_{XY} = \begin{bmatrix} ae & af & be & bf \\ ag & ah & bg & bh \\ ce & cf & de & df \\ cg & ch & dg & dh \end{bmatrix} \tag{5}$$

Given $\rho_{XY}$, it is essential that it should be factored before one can conclude that it represents the product of two quantum variables.

Example 2. Consider $|\varphi\rangle = 0.9|00\rangle - 0.3|01\rangle + 0.1|10\rangle + 0.3|11\rangle$. Its density matrix is:

$$\rho_{XY} = \begin{bmatrix} 0.81 & -0.27 & 0.09 & 0.27 \\ -0.27 & 0.09 & -0.03 & -0.09 \\ 0.09 & -0.03 & 0.01 & 0.03 \\ 0.27 & -0.09 & 0.03 & 0.09 \end{bmatrix} \tag{6}$$

Its representation as product (4) by using some of the conditions of (5) as $\rho_X = \begin{bmatrix} 0.9 & 0 \\ 0 & 0.1 \end{bmatrix}$ and $\rho_Y = \begin{bmatrix} 0.82 & -0.24 \\ -0.24 & 0.18 \end{bmatrix}$ is obviously incorrect since the tensor product of the two does not yield the original density matrix. This points to the possibility that one may inaccurately use a classical representation for the quantum state if the joint state was not known and one was depending on the reconstruction based on (5).

In reality, the probabilities for the pure state of equation (6) are correctly obtained by the decision tree of Figure 4 by the reconstruction outlined at the beginning of this section.

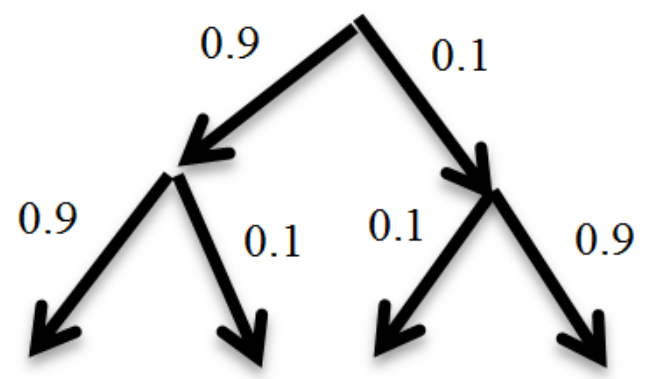

Figure 4. Correct representation of the quantum process

The decision tree of the maximally entangled state $|\varphi\rangle = \frac{1}{\sqrt{2}}(|00\rangle + |11\rangle)$ has different probabilities in the two layers that is sufficient to account for the entanglement as in Figure 5.

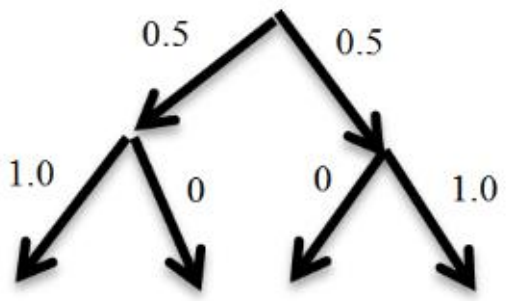

Figure 5. The maximally entangled state

Generalizing from this example, which only requires a flip in the branching probability at the second level, an decision may require finding random decision trees that provide a good approximation to a given quantum state using a minimum of different branching probabilities. This will of course depend on what measure of goodness of approximation is used.

CONSTRAINED TREE APPROXIMATION

We restrict ourselves to two-qubit states. Let the probabilities of 00, 01, 10, and 11 be *A, B, C,* and *D*. These are the values in the diagonal of the density function and *A+B+C+D*=1. We want to constrain the approximation to this case to just two branching probabilities.

The two layered decision tree will be assume to have the branching probabilities of *a, 1-a* in the first layer and *c, 1-c* in the second layer.

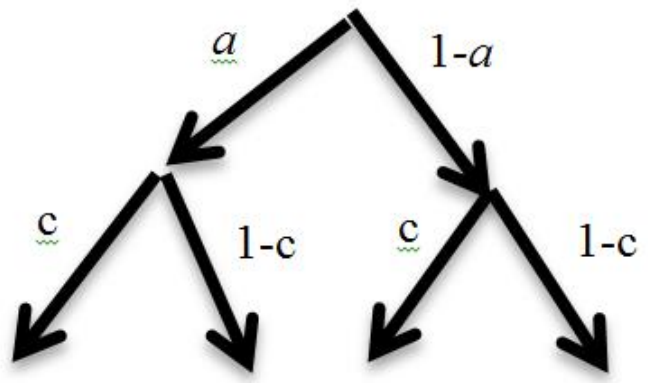

Figure 6. The constrained tree

The following equations are implied:

$$\begin{aligned} ac &\approx A \\ a-ac &\approx B \\ c-ac &\approx C \\ 1-a-ac+ac &\approx D \end{aligned} \qquad (7)$$

The approximation in the relationships comes from the fact that two variables (*a* and *c*) cannot equal the arbitrary values of A,B,C,D. Solving, these equations, we obtain as estimates:

$$\begin{aligned} a &= A+B \\ c &= A+C \end{aligned} \qquad (8)$$

**Theorem 2**. For a quantum state with two independent qubits, the events can be exactly represented by two kinds of branching probabilities, one for each layer.

*Proof*. It follows from a straightforward substitution of the probabilities in a tree.

Example 2. Now consider a source that generates two qubits, where the first qubit is described by the density function $\rho_X = \begin{bmatrix} 0.6 & 0.3 \\ 0.3 & 0.4 \end{bmatrix}$ and the other by $\rho_Y = \begin{bmatrix} 0.7 & 0.1 \\ 0.1 & 0.3 \end{bmatrix}$. Their joint density function will be represented by

$$\rho_{XY} = \begin{bmatrix} 0.42 & 0.06 & 0.21 & 0.03 \\ 0.06 & 0.18 & 0.03 & 0.09 \\ 0.21 & 0.03 & 0.28 & 0.04 \\ 0.03 & 0.09 & 0.04 & 0.12 \end{bmatrix} \quad (9)$$

The probabilities of getting the various combinations of 0s and 1s are:

P(00)=0.42; P(01)=0.18; P(10)= 0.28; and P(11)= 0.12

The branching probabilities a and c are 0.6 and 0.7, respectively. A tree like Figure 7 gives the correct final probabilities.

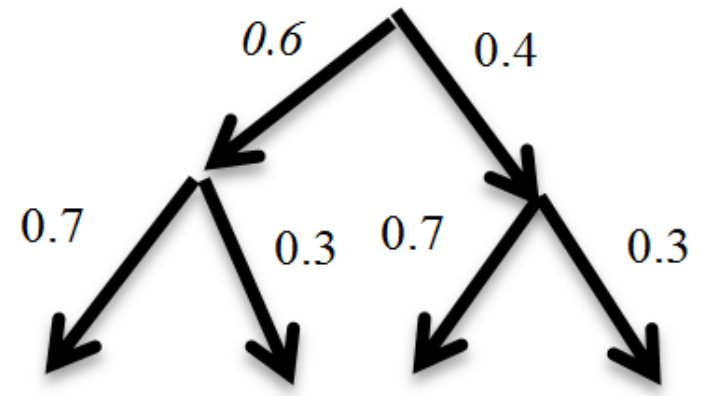

Figure 7. The constrained tree for (9)

If the qubits are not independent then the constrained tree will not be able to exactly replicate the probabilities of the various events in the quantum state.

## DISCUSSION

This paper considered the problem of random tree representations of quantum decision processes. We showed that in the general case where each node can have any branching probability values, such a random tree can accurately represent the probabilities associated with an arbitrary quantum state. Thus one can claim that quantum entanglement is equivalent to different branching probabilities at the decision nodes. For a two-qubit state, we also examined the case of two separate branching probabilities.

Classical agents may ascribe the randomness of probability values at the branching nodes to quantum entanglement or may wish to see it as a purely classical decision process with unexplained correlations in its different components.

REFERENCES

1. Khrennikov, A.Y. Ubiquitous Quantum Structure: From Psychology to Finance. Springer, 2010
2. Busemeyer, J.R. and Bruza, P. Quantum Models of Cognition and Decision. Cambridge University Press, 2012
3. Kak, S. The absent-minded driver problem redux. 2017. arXiv:1702.05778
4. Kak S. The three languages of the brain: quantum, reorganizational, and associative. In Learning as Self-Organization, K. Pribram and J. King (editors). Lawrence Erlbaum Associates, Mahwah, NJ, 185-219, 1996
5. Kak S. Active agents, intelligence, and quantum computing. Information Sciences 2000; 128: 1-17
6. Conte, E., Santacroce, N., Laterza, V., Conte, S., Federici A., Todarello, O. The brain knows more than it admits: A quantum model and its experimental confirmation. Electronic Journal of Theoretical Physics 2012; 9: 72-110
7. Asano, M., Basieva, I., Khrennikov, A., Ohya, M., Yamato, I. 2013. Non-Kolmogorovian Approach to the Context-Dependent Systems Breaking the Classical Probability Law Foundations of Physics 2013; 43: 895-911
8. Kak, A., Gautam, A., and Kak, S. A three-layered model for consciousness states. NeuroQuantology 2016; 14: 166-174
9. Kak, S. Psychophysical parallelism. https://subhask.okstate.edu/sites/default/files/qPsycho.pdf
10. Aerts, D. Quantum structure in cognition. Journal of Mathematical Psychology 2009; 53: 314-348
11. Yukalov, V.I. and Sornette, D. Decision theory with prospect interference and entanglement. Theory and Decision 2010; 70: 283-328
12. Kak, S. The initialization problem in quantum computing. Foundations of Physics 1999; 29: 267-279
13. Kak, S. Quantum information and entropy. International Journal of Theoretical Physics 2007; 46: 860-876
14. Pothos, E. M., and Busemeyer, J. R. Can quantum probability provide a new direction for cognitive modeling. Behavioral and Brain Sciences 2013; 36: 255-274
15. Festinger, L. A theory of cognitive dissonance, volume 2. Stanford university press, 1962.
16. Wicklund, R.A. and Brehm, J.W. Perspectives on cognitive dissonance. Psychology Press, 2013.
17. Asano, M., Basieva, I., Khrennikov, A., Ohya, M., Tanaka, Y. Yamato, I. Quantum-like model for the adaptive dynamics of the genetic regulation of E. coli's metabolism of glucose/lactose. System Synthetic Biology 2012; 6(-2): 1–7

18. Bell, J.S. Speakable and Unspeakable in Quantum Mechanics. Cambridge University Press, 1987.
19. Esfeld, M. Bell's Theorem and the Issue of Determinism and Indeterminism. Foundations of Physics 2015; 45: 471–482
20. Kak, S. Probability constraints and the classical/quantum divide. NeuroQuantology 2013: 1: 600-606
21. Aerts, D. Quantum structure in cognition. Journal of Mathematical Psychology 2009; 53: 314-348
22. Kak, S. Simulating entanglement in classical computing for cryptography applications. Cryptologia 2016; 40: 348-354